\pdfoutput=1
\documentclass[accepted]{safeai2026}
\usepackage[utf8]{inputenc}
\usepackage[american]{babel}

\usepackage{natbib}
\usepackage{mathtools}
\usepackage{booktabs}
\usepackage{tikz}
\usepackage{graphicx}
\usepackage{amsfonts}
\usepackage{amssymb}
\usepackage{bbm}
\usepackage[ruled,vlined]{algorithm2e}
\usepackage{subcaption}
\usepackage{tabularx}
\usepackage{cleveref}
\usepackage{tcolorbox}
\usepackage{xurl}

\newcommand{\mmd}{\text{MMD}}
\newcommand{\hatmmd}{\widehat{\mmd}}

\title{MMD-Flagger: Leveraging Maximum Mean Discrepancy to Detect Hallucinations}

\author[1]{Kensuke Mitsuzawa\thanks{\parbox[t]{\dimexpr\linewidth-2em\relax}{Corresponding author:\\\normalfont kensuke.mitsuzawa@univ-cotedazur.fr}}}
\affil[1]{Université Côte d'Azur, CNRS, LJAD, France}

\author[2]{Damien Garreau}
\affil[2]{Center for Artificial Intelligence and Data Science (CAIDAS),\protect\\ Julius-Maximilans-Universität Würzburg, Germany}

\makeatletter
\renewcommand{\notice@text}{}
\makeatother
\renewpagestyle{plain}{\setfoot{}{\thepage}{}}

\begin{document}
\maketitle

\begin{abstract}
Large Language Models (LLMs) are increasingly integrated into agentic AI systems, yet their propensity to generate hallucinations remains a critical safety concern. 
Detecting these factual errors at test-time, particularly without ground-truth labels, is essential for building trustworthy autonomous agents. 
We propose \textbf{MMD-Flagger}, an hallucination detection method that utilizes Maximum Mean Discrepancy (MMD) and monitors the stability of LLM outputs across varying decoding temperatures. 
Our method tracks the MMD trajectory between a LLM's response at a certain decoding configuration and a set of stochastic samples, identifying hallucinations based on the trajectory's characteristic shape. 
We evaluate MMD-Flagger on multi-lingual claim verification benchmarks (MUCH) using modern LLMs like Llama-3 families and \text{Gemma-3}. 
% Our experiments with MMD-Flagger on these benchmarks show the potential of our method for hallucination detection.
\end{abstract}

\section{Introduction}
\label{sec:introduction}

The rapid deployment of Large Language Models (LLMs) in safety-critical domains such as law, medicine, and autonomous agentic systems is hindered by \textit{hallucinations}, which are fluent but factually incorrect generations \citep{ji_et_al_2023}.
In agentic AI, where models perform sequential decision-making, a single hallucinated step can lead to catastrophic failures~\citep{Cemri-2025-fail, Kim-2025-scaling}.
Ensuring the safety and reliability of these systems requires robust, deployment-time monitoring mechanisms that can audit agent outputs at test-time without relying on external knowledge bases or gold-standard references.

One of the primary drivers of hallucinations is the high sensitivity of LLMs to decoding strategies and parameters \citep{Shi-2024-EMNLP}.
Stochastic decoding paths under different configurations (such as random seeds or temperature shifts) can cause LLM outputs to vary widely.
In consistency-based uncertainty quantification (UQ), a key premise is that when a model is confident and generating grounded, faithful content, its outputs remain semantically consistent across alternative decoding paths \citep{Park-2025-steer}.
Conversely, when the generation contains hallucinations, it is highly sensitive to the decoding configuration, producing unstable and inconsistent semantic meanings.
Therefore, a deployment-time safety monitor can identify hallucinated generations by auditing the stability of a specific prompt-response pair under varying decoding conditions.

Such a safety monitor should operate at test-time without supervision, as ground-truth hallucination labels are unavailable during live deployment.
While existing consistency-based UQ methods estimate generation reliability by comparing multiple output token sequences \citep{Farquhar-2024detecting, Manakul-2023-selfcheckgpt}, they are often restricted to token-level semantic similarities and ignore internal representations.
On the other hand, internal-state and structural UQ methods require supervised training or depend on specific model structures \citep{binkowski-etal-2025-hallucination}.
An unsupervised, non-parametric framework that flexibly monitors generation stability across diverse feature spaces, including internal hidden states, remains an open challenge.

\begin{figure}[t]
    \centering
    \includegraphics[width=\columnwidth]{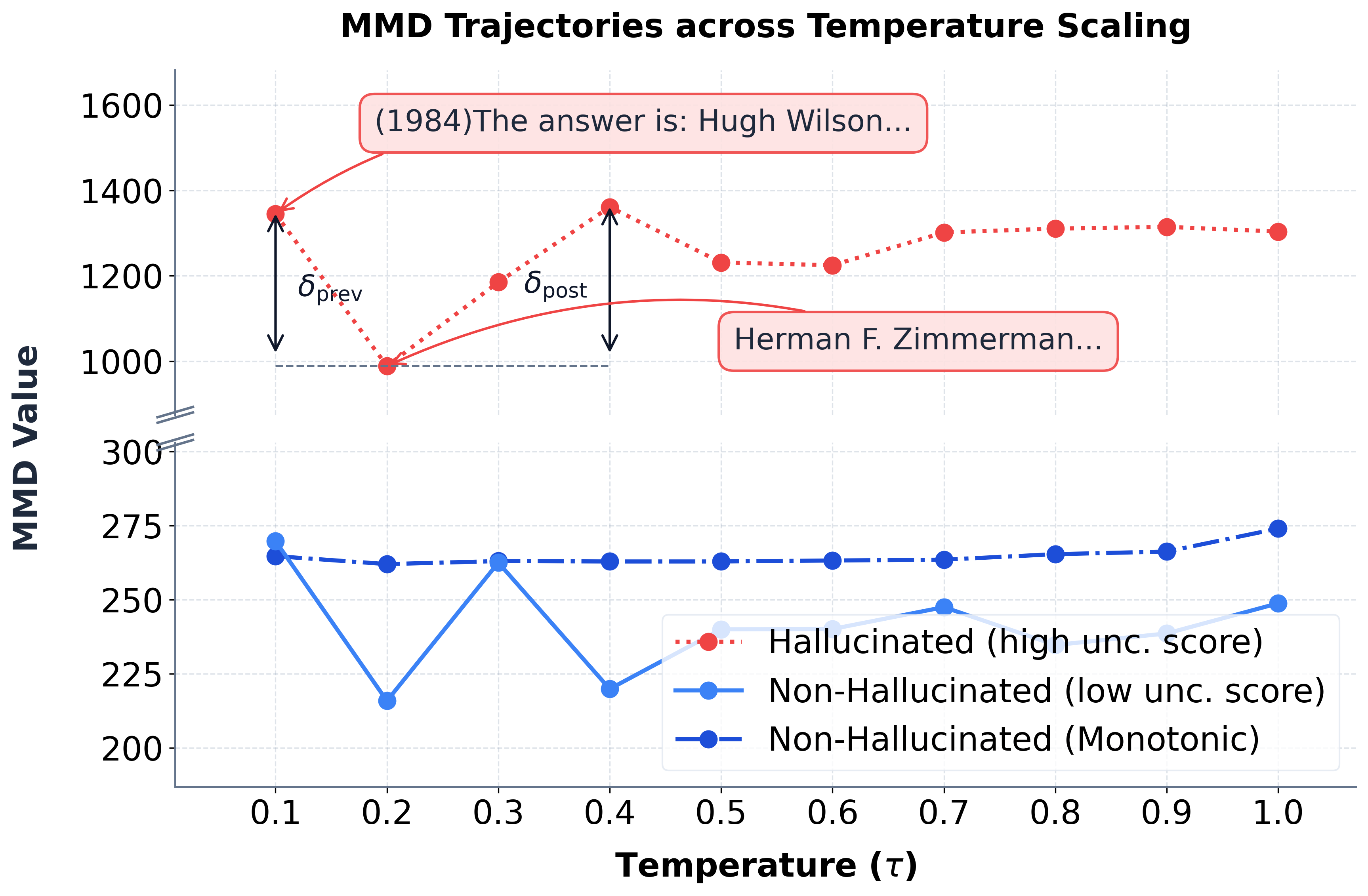}
    \caption{
        Overview of MMD-Flagger.
        % Three lines plot MMD values between generations from the given decoding strategy and stochastic strategies across temperatures~$\tau$.
        Three lines illustrate the trajectories of squared MMD distances between the hypothesis generation and stochastic samples across varying temperatures~$\tau$.
        % Blue lines represent non-hallucinated responses; the red dotted line represents a hallucinated response.
        Blue lines represent the calculated squared MMD values for non-hallucinated responses across temperatures.
        The uncertainty score components $\delta_{\text{prev}}$ and $\delta_{\text{post}}$ (Eq.~\ref{eq:scorer-score}) enlarge as the trajectory forms a sharp U-shape.
        In the hallucinated example, the model responds ``\textit{Herman Wouk Jr.}'' instead of the factually correct answer ``\textit{Hugh Wilson}''.
        Callout boxes display responses generated at each temperature.
    }
    \label{fig:concept-mmd-scorer}
\end{figure}

To address this challenge, we propose \textbf{MMD-Flagger}, an unsupervised, training-free safety monitoring method.
MMD-Flagger measures the stability of a target generation by comparing it to stochastic samples generated under a range of decoding temperatures.
By employing the Maximum Mean Discrepancy (MMD) \citep{Gretton-2012a}, MMD-Flagger constructs a trajectory of MMD values across these temperatures as illustrated in Fig.~\ref{fig:concept-mmd-scorer}.
Our core hypothesis is that high uncertainty manifests as a U-shaped MMD trajectory, where a sharp, non-monotonic dip serves as an active indicator of hallucination, whereas faithful outputs show stable, monotonic MMD growth.

In this paper, we demonstrate that MMD-Flagger is highly flexible and can audit stability across diverse feature spaces such as internal hidden states. 
Furthermore, we present empirical results in hallucination detection across Llama, Gemma model families on the hallucination benchmark dataset.

\section{Related Work}
\label{sec:related-work}

\textbf{Consistency-based uncertainty quantification methods.} 
These methods detect hallucinations by comparing multiple generations. 
\citet{Fomicheva-2020-unsupervised} use Monte-Carlo dropout for uncertainty, while SelfCheckGPT \citep{Manakul-2023-selfcheckgpt} and Semantic Entropy \citep{Nikitin-2024-kernel} rely on external text embedding models or NLI models for comparing candidates against sampled alternatives. 
Meanwhile, MMD-Flagger differs from these methods in the flexibility of features it can utilize.

\textbf{Hallucination detection using LLMs' internal representations.} 
EigenScore \citep{Chen-2024-inside} uses SVD on hidden states to isolate hallucination subspaces. 
LapEigvals \citep{binkowski-etal-2025-hallucination} performs spectral analysis on attention maps. 
Crucially, LapEigvals is a supervised learning approach that requires training a classifier on extracted spectral features to distinguish between grounded and hallucinated content.
% While powerful, these methods can be computationally intensive and may assume specific structural properties that do not always hold across model families.
While powerful, there is a risk that the detection performance depends on the model architecture and the data distribution.

% \textbf{Applications of MMD in various LLM tasks.} 
% In the context of LLM applications, several previous studies have employed MMD.
% For example, \citet{Oblovatny-2025-attention} used MMD to measure distances between prompt and response representations. 
% Unlike their supervised approach using trainable deep kernels, MMD-Flagger is entirely unsupervised and requires no prior training.
% Similarly, \citet{Zhangs-2024MMDMP} proposed a hypothesis testing framework using MMD as a metric.
% The task is to distinguish between human-written and LLM-generated texts.

\begin{table*}[t]
    \centering\small
    \caption{Hallucination Detection Performance on MUCH dataset. The performance score is ROC-AUC. For our proposed MMD-Flagger, results are presented as \textit{Linear} / \textit{Gaussian} kernels. Bold values highlight the best ROC-AUC score per model column.}
    \label{tab:results}
    \setlength{\tabcolsep}{4.5pt}
    \begin{tabular}{lllccc}
    \toprule
    Estimator & Feature & Agg. & \multicolumn{3}{c}{Models} \\
    \cmidrule(lr){4-6}
     & & & Llama-3.1-8B & Llama-3.2-3B & Gemma-3-4B \\
    \midrule
    KLE & Text & -- & .484 & .422 & .568 \\
    % NumSets & Text & -- & \textbf{.629} & .470 & .425 \\
    % EigV & Text & -- & .507 & .586 & .482 \\
    % Deg & Text & -- & .538 & .494 & .395 \\
    Ecc & Text & -- & .529 & .569 & .525 \\
    LexSim & Text & -- & .460 & .503 & .392 \\
    EigenScore & Hidden states & last & .549 & .398 & .579 \\
      &   & avg & .326 & .506 & .537 \\
    \midrule
    \textbf{MMD-Flagger (ours)}
     & Word embedding & avg & .534 / .515 & .620 / \textbf{.637} & .398 / .442 \\
     & Hidden states & last & .555 / .452 & .473 / .569 & .362 / .500 \\
     &   & avg & .508 / .446 & .567 / .551 & .408 / .500 \\
    %  & Text embedding & -- & .584 / .480 & .589 / .472 & .359 / .522 \\
     & LapEigvals & top-10 & .586 / .500 & .480 / .620 & .742 / .471 \\
     &   & top-100 & .622 / .500 & .469 / .562 & .558 / .500 \\
     & AttentionMat & top-10 & .529 / .500 & .441 / .562 & .498 / .471 \\
     &   & top-100 & .509 / .500 & .457 / .562 & .394 / .500 \\
     & Ensemble, Sec~\ref{sec:mmd-flagger-ensemble} & -- & \textbf{.624} & .516 & \textbf{.754} \\
    \bottomrule
    \end{tabular}
\end{table*}

\section{Methodology}
\label{sec:methodology}

\subsection{Stochastic Sampling and Temperature}

% topic: definition of LLM generation, features and decoding strategy (including its parameters).
We present the minimal formulation used in MMD-Flagger.
We define the LLM's generation process as $\mathbf{h} = \text{LLM}(x; \theta)$, where $x$ is the prompt, $\theta$ represents the decoding strategy and its parameters, and $\mathbf{h}$ is the feature representation of the generated text.
The feature representation $\mathbf{h}$ can be an arbitrary $d$-dimensional embedding vector, such as token embeddings or hidden states, or a more complex representation derived from the attention matrix~\citep{binkowski-etal-2025-hallucination}.
% topic: definition for the hypothesis.
We define $\mathbf{h}_\text{hyp}$ as the representation of the output under a selected decoding strategy $\theta_\text{hyp}$, where $\theta_\text{hyp}$ is determined by the user or calibration algorithms (i.e., \citep{Yang-2024-QuATS, Xie-2024-ATS, Dang-2026-TAMPO}).
% topic: definition for the stochastic generation.
We denote the stochastic decoding strategy at temperature $\tau \in \mathbb{R}$ as~$\theta_{\text{sto}}^{\tau}$, and the resulting extracted feature as $\mathbf{h}_{\text{sto}}^{\tau}$.

% topic: definition of sample sets.
By iterating the generation process multiple times, we obtain two sets of samples, \mbox{$S_{\text{hyp}} := \{\mathbf{h}_{\text{hyp}}^{1}, \dots, \mathbf{h}_{\text{hyp}}^{n} \}$} and~\mbox{$S_{\text{sto}}^{\tau} := \{ \mathbf{h}_{\text{sto}}^{\tau, 1}, \dots, \mathbf{h}_{\text{sto}}^{\tau, m} \}$}, where $n$ and $m$ denote the number of samples in each set.

\subsection{Maximum Mean Discrepancy (MMD)}

MMD~\citep{Gretton-2012a} measures the distance between two probability distributions $P$ and $Q$ by mapping them into a Reproducing Kernel Hilbert Space (RKHS). 
We define $k(\cdot, \cdot)$ as a positive-definite kernel function mapping the feature vectors to a Reproducing Kernel Hilbert Space (RKHS)~\citep{Fukumizu-2007}.
The squared MMD is defined as:
\begin{equation*}
  \begin{split}
    \mmd_k^2(P, Q) := & \mathbb{E}_{A, A' \sim P}[k(A, A')] \\
    & + \mathbb{E}_{B, B' \sim Q}[k(B, B')] \\
    & - 2\mathbb{E}_{A \sim P, B \sim Q}[k(A, B)].
  \end{split}
\end{equation*}

In the setting of MMD-Flagger, the sample sets $A$ and $B$ are~\mbox{$S_{\text{hyp}}$} and $S_{\text{sto}}^{\tau}$, respectively.
$P$ and $Q$ represent the probability distributions of generating the feature vector $\mathbf{h}$ that represents an LLM's generation process for a given prompt~$x$.
Given these two sample sets, the unbiased quadratic estimator of MMD is defined as
\begingroup
\small
\begin{equation} 
  \label{eq:mmd-unbiased-est}
  \begin{split}
    \hatmmd^{2}_U & ( S_{\text{hyp}}, S_{\text{sto}}^{\tau} ) := \\
    & \frac{1}{n (n-1)} \sum_{1 \leq i \ne i' \leq n} k(\mathbf{h}_{\text{hyp}}^{i}, \mathbf{h}_{\text{hyp}}^{i'}) \\
    & + \frac{1}{m(m-1)} \sum_{1 \leq j \ne j' \leq m} k(\mathbf{h}_{\text{sto}}^{\tau, j}, \mathbf{h}_{\text{sto}}^{\tau, j'}) \\
    & - \frac{2}{nm} \sum_{i=1}^n \sum_{j=1}^m k(\mathbf{h}_{\text{hyp}}^{i}, \mathbf{h}_{\text{sto}}^{\tau, j}).
  \end{split}
\end{equation}
\endgroup

The choice of the kernel function is arbitrary, and this work evaluates two specific kernels. 
The first is the Gaussian kernel, $k^{\text{Gau}}(\mathbf{h}_{\text{hyp}}, \mathbf{h}_{\text{sto}}^{\tau}) = \exp(-\|\mathbf{h}_{\text{hyp}}-\mathbf{h}_{\text{sto}}^{\tau}\|^2 / 2\gamma^2)$, where the bandwidth $\gamma$ is selected as the 25-th percentile of pairwise distances in a calibration set~\citep{Garreau2018-largesampleanalysismedian}.
The second is the linear kernel, $k^{\text{lin}}(\mathbf{h}_{\text{hyp}}, \mathbf{h}_{\text{sto}}^{\tau}) = \mathbf{h}_{\text{hyp}}^\top \mathbf{h}_{\text{sto}}^{\tau}$.

In some cases, the sample set $S_{\text{hyp}}$ can consist of a single LLM generation (i.e., $n=1$).
Appendix~\ref{sec:case-one-sample-generation} provides the definition of MMD for such cases.

\subsection{MMD-Flagger Algorithm}

The MMD-Flagger algorithm (Algorithm~\ref{alg:mmd-scorer}) computes an uncertainty score $\delta$ based on the non-monotonicity of the MMD trajectory $D(\theta_{\text{hyp}}) = (\hatmmd_U^2(S_{\text{hyp}}, S_{\text{sto}}^{\tau}))_{\tau \in T}$ as a function of a decoding strategy.
This approach is grounded in the assumption that \textit{faithful} LLM generations exhibit semantic consistency across sampling temperatures.
For faithful outputs, the MMD trajectory $D(\theta_{\text{hyp}})$ typically exhibits either monotonic growth due to increasing stochastic diversity or a shallow U-shape (Fig~\ref{fig:concept-mmd-scorer}).
Conversely, unfaithful outputs produce a deeper, more acute U-shape.
Consequently, MMD-Flagger assigns higher uncertainty scores~$\delta$ to deeper and more pronounced U-shapes.

On the MMD trajectory $D(\theta_{\text{hyp}})$, we define $\delta_{\mathrm{prev}}$ and $\delta_{\mathrm{post}}$ as the absolute difference between the minimum MMD value and the local maximum MMD values at the nearest temperatures, respectively (Fig~\ref{fig:concept-mmd-scorer}):
\begingroup
\small
\begin{align}
  \delta_{\mathrm{prev}} &= \Big| \mathop{\mathrm{argmin}}_{\tau \in T} D(\tau) - \mathop{\mathrm{argmax}}_{\tau < \tau_{\mathrm{argmin}}} D(\tau) \Big|, \label{eq:delta-prev} \\
  \delta_{\mathrm{post}} &= \Big| \mathop{\mathrm{argmin}}_{\tau \in T} D(\tau) - \mathop{\mathrm{argmax}}_{\tau > \tau_{\mathrm{argmin}}} D(\tau) \Big|, \label{eq:delta-post}
\end{align}
\endgroup
where $\tau_{\mathrm{argmin}} = \arg\min_{\tau \in T} D(\tau)$.
With two terms, the uncertainty score $\delta$ is defined as
\begin{equation}
\label{eq:scorer-score}
\delta = \delta_{\mathrm{prev}} + \delta_{\mathrm{post}}.
\end{equation}

\subsection{MMD-Flagger (Ensemble)}
\label{sec:mmd-flagger-ensemble}

MMD-Flagger is flexible in that it can be applied to different feature representations, such as word embeddings or hidden states.
To systematically exploit the complementary features of generative trajectories across multiple semantic scales, we propose MMD-Flagger (Ensemble).
We denote~\mbox{$\mathbf{D}_d = (\hatmmd_U^2(S_{\text{hyp}, d}, S_{\text{sto}, d}^{\tau}))_{\tau \in T}$} as the MMD trajectory for each feature domain $d \in \mathcal{D}$.

Not all feature domains $\mathcal{D}$ capture useful semantic information.
To filter out inactive feature domains, MMD-Flagger (Ensemble) computes the standard deviation of each trajectory:
$\sigma(\mathbf{D}_d) = \sqrt{\frac{1}{|T|} \sum_{\tau \in T} \left( \mathbf{D}_d(\tau) - \mu(\mathbf{D}_d) \right)^2},$
where~$\mu(\mathbf{D}_d) = \frac{1}{|T|} \sum_{\tau \in T} \mathbf{D}_d(\tau)$.
If $\sigma(\mathbf{D}_d) < \eta$ with threshold $\eta = 10^{-5}$, domain $d$ is discarded, leaving active domains $\mathcal{D}_{\text{active}} \subseteq \mathcal{D}$.

For each active domain $d \in \mathcal{D}_{\text{active}}$, we compute two metrics, the Coefficient of Variation ($\text{CV}_d$) and the entropy~($\mathcal{E}_d$).
The $\text{CV}_d$ is a scale-invariant metric and detects variance spikes independent of representation magnitude~\citep{groenendijk2021multi}, defined as
$\text{CV}_d = \frac{\sigma(\mathbf{D}_d)}{\mu(\mathbf{D}_d) + \epsilon},$ where~$\epsilon = 10^{-8}$ is a numerical stabilizer.
Furthermore, we employ an entropy-based heuristic, $\mathcal{E}_d$, to capture the peakiness (sparsity of fluctuations) within the MMD trajectory.
First, we compute the pairwise distance matrix~\mbox{$\mathbf{M}_d \in \mathbb{R}^{|T| \times |T|}$}, where $\mathbf{M}_d(i, j) = | \mathbf{D}_d(\tau_i) - \mathbf{D}_d(\tau_j) |$, to represent the volatility of the trajectory across temperatures.
Then, $\mathbf{M}_d$ is flattened and normalized to a probability distribution $\mathbf{p}_d(k) = \frac{\mathbf{M}_d(k)}{\sum_{l} \mathbf{M}_d(l)}$, where $k$ indexes the elements of the flattened matrix.
Using $\mathbf{p}_d(k)$, we compute the relative entropy as \mbox{$\mathcal{E}_d = 1.0 - \frac{\text{H}(\mathbf{p}_d)}{\log(|T|^2) + \epsilon}$}, where~\mbox{$\text{H}(\mathbf{p}_d) = -\sum_{k} \mathbf{p}_d(k) \log \mathbf{p}_d(k)$}.
The uncertainty score for $d \in \mathcal{D}$ is computed as $s_d = \text{CV}_d + 2.0 \times \mathcal{E}_d$.

The final ensemble uncertainty score $\delta_{\text{ensemble}}$ is computed as the consensus mean across all active domains:
\begin{equation}
  \label{eq:ensemble-score}
  \delta_{\text{ensemble}} = \frac{1}{|\mathcal{D}_{\text{active}}|} \sum_{d \in \mathcal{D}_{\text{active}}} s_d.
\end{equation}
If no active domains are detected, $\delta_{\text{ensemble}} = 0$.

\begin{algorithm}[t]
    \small
    \caption{\label{alg:mmd-scorer}MMD-Flagger}
    \SetAlgoLined
    \KwIn{LLM, a prompt $x$, a decoding strategy $\theta_{\text{hyp}}$, a set of temperatures $T$}
    \KwOut{Uncertainty score $\delta$}
    Generate $S_{\text{hyp}} = \{ \text{LLM}(x; \theta_{\text{hyp}}) \}_{i=1}^n$\;
    \ForEach{$\tau \in T$}{
        Sample $S_{\text{sto}}^{\tau} = \{ \text{LLM}(x; \theta_{\text{sto}}^{\tau}) \}_{i=1}^m$\;
    }
    Compute $D(\theta_{\text{hyp}}) = (\hatmmd_U^2(S_{\text{hyp}}, S_{\text{sto}}^{\tau}))_{\tau \in T}$ via \eqref{eq:mmd-unbiased-est}\;
    Compute $\delta$ via \eqref{eq:scorer-score}. \\
    \Return{$\delta$}\;
\end{algorithm}

\section{Empirical Assessment}
\label{sec:assessment}

% \subsection{Benchmarks and Models}

% dataset
We evaluate MMD-Flagger on the multilingual hallucination benchmark dataset,~\textit{MUCH}~\citep{dentan2025much}.
We use models: \textit{Llama-3.1-8B}, \textit{Llama-3.2-3B}, \textit{Gemma-3-4B}.

% Estimators
\textbf{Estimators}: We compare the performance of the hallucination detection task with the following baselines.
\textit{KLE}~\citep{Nikitin-2024-kernel} and \textit{Ecc}~\citep{Lin-2024-generating} are black-box estimators based on sampling responses by leveraging external text embedding models or NLI models.
\textit{LexSim}~\citep{Fomicheva-2020-unsupervised} is a black-box estimator that measures the average surface lexical overlap (i.e., ROUGE or Jaccard similarity).
\textit{EigenScore}~\citep{Chen-2024-inside} is a white-box estimator that measures the stability of the internal features of LLMs (the middle layer).

% Features
For MMD-Flagger, we set the following features.
\texttt{Word embedding} is a feature vector without positional information, obtained at the first layer of the model.
\texttt{Hidden states} is a feature vector extracted from the middle hidden layer of each model, following the setting of \citep{Chen-2024-inside}.
We form two types of features from the hidden states: \texttt{last token} is a feature vector of the last token, and \texttt{avg} is the mean of the hidden states over all tokens.
\texttt{LapEigVals}~\mbox{\citep{binkowski-etal-2025-hallucination}} is a customized feature vector of taking top-k eigenvalues of Laplacians derived from attention matrices. 
\texttt{AttentionMat}~\mbox{\citep{Chuang-etal-2024-lookback}} is a feature vector of raw attention maps without Laplacian.  
% \texttt{Text embedding} is a feature vector obtained from the external model \textbf{DeBERTa} that is the same setting as \textit{KLE}.

% Sampling conditions
The MUCH dataset defines the default decoding strategy $\theta_{\text{hyp}}$.
For the stochastic generation $\theta_{\text{sto}}^\tau$, we set~\mbox{$T = \{0.1, 0.2, ..., 1.0\}$}, and $m=5$ for each $\tau$.

\subsection{Results}

As shown in Table~\ref{tab:results}, the optimal hallucination detection estimator is highly model-dependent, likely driven by differences in LLM architectures.
% For example, given the strong performance of text-based approaches on Ministral-8B, hallucination-related signals for certain architectures may reside primarily in the final layers or logits.

Overall, the empirical results strongly demonstrate the effectiveness of our proposed framework.
Specifically, MMD-Flagger (Ensemble) achieves the best performance on Llama-3.1-8B and Gemma-3-4B, while the single feature achieves the best detection at Llama-3.2-3B.
That may indicate the necessity of more sophisticated feature selection mechanism or ensembling strategy for certain model architectures.

\section{Limitations}

Like other consistency-based methods, MMD-Flagger requires multiple stochastic generations ($50$ in this study) per prompt, which can be computationally expensive.
Future work will explore reducing this sampling overhead to improve applicability for larger models and real-time settings.

\section{Conclusion}
\label{sec:conclusion}

In this paper, we propose MMD-Flagger, a test-time hallucination detection method using Maximum Mean Discrepancy trajectories across decoding temperatures.
Our evaluations show potentially promising results on the Llama and Gemma-3 families.
Future work will evaluate broader architectures and datasets to enhance generalizability.

\begin{contributions}
    The first author designed the MMD-Flagger framework and conducted the experiments.
    The second author provided theoretical guidance and supervised the benchmark evaluations.
\end{contributions}

\begin{acknowledgements}
%     This work was supported by the Safe AI Research Grant.
%     We thank the reviewers for their constructive feedback.
    This work has been supported by the French government, through the NIM-ML project (ANR-21-CE23-0005-01). 
    This work was performed using HPC resources from GENCI-IDRIS (Grant 2025-A0191016865).
    We extend our sincere thanks to Prof. Motonobu Kanagawa for insightful discussions and valuable feedbacks on the conceptualization of this work.
\end{acknowledgements}

\bibliography{main}

\begin{thebibliography}{24}
\providecommand{\natexlab}[1]{#1}
\providecommand{\url}[1]{\texttt{#1}}
\expandafter\ifx\csname urlstyle\endcsname\relax
  \providecommand{\doi}[1]{doi: #1}\else
  \providecommand{\doi}{doi: \begingroup \urlstyle{rm}\Url}\fi

\bibitem[Binkowski et~al.(2025)Binkowski, Janiak, Sawczyn, Gabrys, and
  Kajdanowicz]{binkowski-etal-2025-hallucination}
Jakub Binkowski, Denis Janiak, Albert Sawczyn, Bogdan Gabrys, and Tomasz~Jan
  Kajdanowicz.
\newblock Hallucination detection in {LLM}s using spectral features of
  attention maps.
\newblock In \emph{Proceedings of the 2025 Conference on Empirical Methods in
  Natural Language Processing}, pages 24354--24385, 2025.

\bibitem[Cemri et~al.(2025)Cemri, Pan, Yang, Agrawal, Chopra, Tiwari, Keutzer,
  Parameswaran, Klein, Ramchandran, Zaharia, Gonzalez, and
  Stoica]{Cemri-2025-fail}
Mert Cemri, Melissa~Z. Pan, Shuyi Yang, Lakshya~A. Agrawal, Bhavya Chopra,
  Rishabh Tiwari, Kurt Keutzer, Aditya Parameswaran, Dan Klein, Kannan
  Ramchandran, Matei Zaharia, Joseph~E. Gonzalez, and Ion Stoica.
\newblock {Why Do Multi-Agent LLM Systems Fail?}
\newblock In \emph{Advances in Neural Information Processing Systems}, 2025.

\bibitem[Chen et~al.(2024)Chen, Liu, Chen, Gu, Wu, Tao, Fu, and
  Ye]{Chen-2024-inside}
Chao Chen, Kai Liu, Ze~Chen, Yi~Gu, Yue Wu, Mingyuan Tao, Zhihang Fu, and
  Jieping Ye.
\newblock {INSIDE}: {LLM}s' internal states retain the power of hallucination
  detection.
\newblock In \emph{The Twelfth International Conference on Learning
  Representations}, 2024.

\bibitem[Chuang et~al.(2024)Chuang, Qiu, Hsieh, Krishna, Kim, and
  Glass]{Chuang-etal-2024-lookback}
Yung-Sung Chuang, Linlu Qiu, Cheng-Yu Hsieh, Ranjay Krishna, Yoon Kim, and
  James~R. Glass.
\newblock Lookback lens: Detecting and mitigating contextual hallucinations in
  large language models using only attention maps.
\newblock In \emph{Proceedings of the 2024 Conference on Empirical Methods in
  Natural Language Processing}, pages 1419--1436, 2024.

\bibitem[Dang et~al.(2026)Dang, Lan, Wan, Zhao, and Lu]{Dang-2026-TAMPO}
Haoran Dang, Cuiling Lan, Hai Wan, Xibin Zhao, and Yan Lu.
\newblock {Temperature as a Meta-Policy: Adaptive Temperature in {LLM}
  Reinforcement Learning}.
\newblock In \emph{International Conference on Learning Representations}, 2026.

\bibitem[Dentan et~al.(2025)Dentan, Canesse, Buscaldi, Shabou, and
  Vanier]{dentan2025much}
J{\'e}r{\'e}mie Dentan, Alexi Canesse, Davide Buscaldi, Aymen Shabou, and Sonia
  Vanier.
\newblock {MUCH}: A multilingual claim hallucination benchmark.
\newblock \emph{arXiv preprint arXiv:2511.17081}, 2025.

\bibitem[Farquhar et~al.(2024)Farquhar, Kossen, Kuhn, and
  Gal]{Farquhar-2024detecting}
Sebastian Farquhar, Jannik Kossen, Lorenz Kuhn, and Yarin Gal.
\newblock {Detecting hallucinations in large language models using semantic
  entropy}.
\newblock \emph{Nature}, 630\penalty0 (8017):\penalty0 625--630, 2024.

\bibitem[Fomicheva et~al.(2020)Fomicheva, Sun, Yankovskaya, Blain, Guzm{\'a}n,
  Fishel, Aletras, Chaudhary, and Specia]{Fomicheva-2020-unsupervised}
Marina Fomicheva, Shuo Sun, Lisa Yankovskaya, Fr{\'e}d{\'e}ric Blain, Francisco
  Guzm{\'a}n, Mark Fishel, Nikolaos Aletras, Vishrav Chaudhary, and Lucia
  Specia.
\newblock {Unsupervised quality estimation for neural machine translation}.
\newblock \emph{Transactions of the Association for Computational Linguistics},
  8:\penalty0 539--555, 2020.

\bibitem[Fukumizu et~al.(2007)Fukumizu, Gretton, Sun, and
  Sch\"{o}lkopf]{Fukumizu-2007}
Kenji Fukumizu, Arthur Gretton, Xiaohai Sun, and Bernhard Sch\"{o}lkopf.
\newblock {Kernel Measures of Conditional Dependence}.
\newblock In \emph{Advances in Neural Information Processing Systems},
  volume~20. Curran Associates, Inc., 2007.
\newblock % Tags: core-works, MMD.

\bibitem[Garreau et~al.(2017)Garreau, Jitkrittum, and
  Kanagawa]{Garreau2018-largesampleanalysismedian}
Damien Garreau, Wittawat Jitkrittum, and Motonobu Kanagawa.
\newblock Large sample analysis of the median heuristic.
\newblock \emph{arXiv preprint arXiv:1707.07269}, 2017.

\bibitem[Gretton et~al.(2012)Gretton, Borgwardt, Rasch, Sch{{\"o}}lkopf, and
  Smola]{Gretton-2012a}
Arthur Gretton, Karsten~M. Borgwardt, Malte~J. Rasch, Bernhard Sch{{\"o}}lkopf,
  and Alexander Smola.
\newblock {A Kernel Two-Sample Test}.
\newblock \emph{Journal of Machine Learning Research}, 13\penalty0
  (25):\penalty0 723--773, 2012.
\newblock % Tags: core-works, MMD.

\bibitem[Groenendijk et~al.(2021)Groenendijk, Karaoglu, Gevers, and
  Mensink]{groenendijk2021multi}
Rick Groenendijk, Sezer Karaoglu, Theo Gevers, and Thomas Mensink.
\newblock Multi-loss weighting with coefficient of variations.
\newblock In \emph{Proceedings of the IEEE/CVF Winter Conference on
  Applications of Computer Vision (WACV)}, pages 1468--1477, 2021.

\bibitem[Ji et~al.(2023)Ji, Lee, Frieske, Yu, Su, Xu, Ishii, Bang, Madotto, and
  Fung]{ji_et_al_2023}
Ziwei Ji, Nayeon Lee, Rita Frieske, Tiezheng Yu, Dan Su, Yan Xu, Etsuko Ishii,
  Ye~Jin Bang, Andrea Madotto, and Pascale Fung.
\newblock {Survey of hallucination in natural language generation}.
\newblock \emph{ACM Computing Surveys}, 55\penalty0 (12):\penalty0 1--38, 2023.

\bibitem[Kim et~al.(2025)Kim, Gu, Park, Park, Schmidgall, Heydari, Yan, Zhang,
  Zhuang, Malhotra, et~al.]{Kim-2025-scaling}
Yubin Kim, Ken Gu, Chanwoo Park, Chunjong Park, Samuel Schmidgall, A~Ali
  Heydari, Yao Yan, Zhihan Zhang, Yuchen Zhuang, Mark Malhotra, et~al.
\newblock {Towards a Science of Scaling Agent Systems}.
\newblock \emph{arXiv preprint arXiv:2512.08296}, 2025.

\bibitem[Kornblith et~al.(2019)Kornblith, Norouzi, Lee, and
  Hinton]{Kornblith-2019-similarity}
Simon Kornblith, Mohammad Norouzi, Honglak Lee, and Geoffrey Hinton.
\newblock Similarity of neural network representations revisited.
\newblock In \emph{Proceedings of the 36th International Conference on Machine
  Learning}, volume~97 of \emph{Proceedings of Machine Learning Research},
  pages 3519--3529. PMLR, 2019.

\bibitem[Lin et~al.(2024)Lin, Trivedi, and Sun]{Lin-2024-generating}
Zhen Lin, Shubhendu Trivedi, and Jimeng Sun.
\newblock {Generating with Confidence: Uncertainty Quantification for Black-box
  Large Language Models}.
\newblock \emph{Transactions on Machine Learning Research}, 2024.

\bibitem[Manakul et~al.(2023)Manakul, Liusie, and
  Gales]{Manakul-2023-selfcheckgpt}
Potsawee Manakul, Adian Liusie, and Mark Gales.
\newblock {S}elf{C}heck{GPT}: Zero-resource black-box hallucination detection
  for generative large language models.
\newblock In \emph{Proceedings of the 2023 Conference on Empirical Methods in
  Natural Language Processing}, 2023.

\bibitem[Muandet et~al.(2017)Muandet, Fukumizu, Sriperumbudur, and
  Sch\"{o}lkopf]{Muandet-2017}
K.~Muandet, K.~Fukumizu, B.~K. Sriperumbudur, and B.~Sch\"{o}lkopf.
\newblock {Kernel Mean Embedding of Distributions: A Review and Beyond}.
\newblock \emph{Foundations and Trends in Machine Learning}, 10\penalty0
  (1--2):\penalty0 1--141, 2017.

\bibitem[Nikitin et~al.(2024)Nikitin, Kossen, Gal, and
  Marttinen]{Nikitin-2024-kernel}
Alexander~V. Nikitin, Jannik Kossen, Yarin Gal, and Pekka Marttinen.
\newblock {Kernel Language Entropy: Fine-grained Uncertainty Quantification for
  {LLM}s from Semantic Similarities}.
\newblock In \emph{The Thirty-eighth Annual Conference on Neural Information
  Processing Systems}, 2024.

\bibitem[Park et~al.(2025)Park, Du, Yeh, Wang, and Li]{Park-2025-steer}
Seongheon Park, Xuefeng Du, Min-Hsuan Yeh, Haobo Wang, and Yixuan Li.
\newblock Steer {LLM} latents for hallucination detection.
\newblock In \emph{Forty-second International Conference on Machine Learning},
  2025.

\bibitem[Shi et~al.(2024)Shi, Yang, Cai, Zhang, Wang, Yang, and
  Lam]{Shi-2024-EMNLP}
Chufan Shi, Haoran Yang, Deng Cai, Zhisong Zhang, Yifan Wang, Yujiu Yang, and
  Wai Lam.
\newblock A thorough examination of decoding methods in the era of {LLM}s.
\newblock In \emph{Proceedings of the 2024 Conference on Empirical Methods in
  Natural Language Processing}, pages 8601--8629, 2024.

\bibitem[Vashurin et~al.(2025)Vashurin, Fadeeva, Vazhentsev, Rvanova, Vasilev,
  Tsvigun, Petrakov, Xing, Sadallah, Grishchenkov, Panchenko, Baldwin, Nakov,
  Panov, and Shelmanov]{Shelmanovvashurin-2025}
Roman Vashurin, Ekaterina Fadeeva, Artem Vazhentsev, Lyudmila Rvanova, Daniil
  Vasilev, Akim Tsvigun, Sergey Petrakov, Rui Xing, Abdelrahman Sadallah,
  Kirill Grishchenkov, Alexander Panchenko, Timothy Baldwin, Preslav Nakov,
  Maxim Panov, and Artem Shelmanov.
\newblock Benchmarking uncertainty quantification methods for large language
  models with lm-polygraph.
\newblock \emph{Transactions of the Association for Computational Linguistics},
  13:\penalty0 220--248, 03 2025.

\bibitem[Xie et~al.(2024)Xie, Chen, Lee, Mitchell, and Finn]{Xie-2024-ATS}
Jordan Xie, A.~S. Chen, Y.~Lee, E.~Mitchell, and C.~Finn.
\newblock {Calibrating Language Models with Adaptive Temperature Scaling}.
\newblock In \emph{Proceedings of the 2024 Conference on Empirical Methods in
  Natural Language Processing}, 2024.

\bibitem[Yang and Zhao(2024)]{Yang-2024-QuATS}
Dingming Yang and Hai Zhao.
\newblock {Are {LLM}s Aware that Some Questions are not Open-ended?}
\newblock In \emph{Proceedings of the 2024 Conference on Empirical Methods in
  Natural Language Processing}, pages 8630--8645, 2024.

\end{thebibliography}

% Appendices
\appendix

\section{Assessment Settings and Details}
\label{sec:assessment-settings}

This appendix provides the comprehensive settings, parameters, and prompts required to reproduce our empirical assessment.
We detail the benchmark dataset, model configurations, feature extraction methods, baseline implementations, and the specific parameters used for MMD-Flagger.

\subsection{Dataset and Prompts}

We evaluate all methods on the \texttt{test} split of the MUCH (Multilingual Claim Hallucination) benchmark dataset~\citep{dentan2025much}\footnote{The dataset is downloaded from the official HuggingFace Datasets repository at \url{https://huggingface.co/datasets/orailix/MUCH}.}.
Figure~\ref{fig:much-generation-prompt} shows the exact prompt template used for the generation of the MUCH dataset.

\begin{figure}[htbp]
    \centering
    \begin{tcolorbox}[
        colback=blue!3!white,
        colframe=blue!50!black,
        title={Prompt used for the generation of MUCH dataset},
        fonttitle=\bfseries\sffamily,
        coltitle=white,
        arc=4pt,
        boxrule=1.2pt,
        left=10pt,
        right=10pt,
        top=8pt,
        bottom=8pt
    ]
        \textbf{System:} You are a helpful assistant.
        Always answer questions directly.
        Your answers should be very concise and precise. \\
        \textbf{User:} \texttt{<question>} \\
        \textbf{Assistant:}
    \end{tcolorbox}
    \caption{Generation prompt used for the MUCH dataset.}
    \label{fig:much-generation-prompt}
\end{figure}

The MUCH dataset gives the hallucination label for each prompt and response.
The original hallucination labels are given to sub-phrases in a response.
We compute the ratio of the hallucination per response and regard it as the hallucinated when the ratio is greater than $0.5$.
% We limit the evaluation on the test split of the dataset since this split is manually annotated by human, while the training split is automatically labeled by LLM-as-a-judge.

\subsection{Evaluated Models}

We perform our assessment across large language models (LLMs) downloaded from HuggingFace:
\begin{itemize}
    % \item \texttt{mistralai/Ministral-8B-Instruct-2410}
    \item \texttt{meta-llama/Llama-3.1-8B-Instruct}
    \item \texttt{meta-llama/Llama-3.2-3B-Instruct}
    \item \texttt{google/gemma-3-4b-it}
\end{itemize}

\subsection{Decoding Parameters \texorpdfstring{$\theta_{\text{hyp}}$ and $\theta_{\text{sto}}$}{theta\_hyp and theta\_sto}}

The MUCH dataset defines the decoding strategy and parameters for each generation\footnote{The parameter settings are available at \url{https://huggingface.co/datasets/orailix/MUCH-configs}.}.
We regard these parameters as $\theta_{\text{hyp}}$.
Since the MUCH dataset contains the generated responses for each model, prompt, and decoding parameters, we regard these responses as the features, setting $n=1$.

For the stochastic generation, we set~\mbox{$T = \{0.1, 0.2, ..., 0.9, 1.0\}$}, and sample $m = 5$ responses for each $\tau$ parameter.

\subsection{Feature Extraction and Representation}

We evaluate four distinct feature types extracted from the models.

\subsubsection{Hidden States}

We extract features from the middle hidden layer of each model.
Specifically, for a model with a total of $L$ layers, we extract the hidden states at layer $L/2$.

For forming a feature vector from hidden states, we aggregate the hidden state vectors using \texttt{mean} and \texttt{last-token}.
The aggregation \texttt{mean} takes all tokens in the generated response, and \texttt{last-token} takes the last token in the generated response.

\subsubsection{LapEigvals}

The spectral features of attention maps are computed using the LapEigvals method~\citep{binkowski-etal-2025-hallucination}.\footnote{The source code for this implementation is obtained from \url{https://github.com/graphml-lab-pwr/lapeigvals}.}
LapEigvals computes a normalized laplacian matrix for an attention matrix for each layer and head.
The method extracts the top-$k$ eigenvalues for each normalized Laplacian matrix and concatenates them to form a single feature vector. 
Thus, the dimension of this feature is~\mbox{$L \times H \times \mathbb{R}^k$}, where~$L$ is the number of layers, $H$ is the number of heads, and $k$ is the number of eigenvalues.
We set $k=\{10, 100\}$.
While the original paper applies PCA (Principal Component Analysis) to reduce the dimension, we do not apply PCA to the eigenvalues to preserve the high-dimensional spectral features, considering that MMD can handle high-dimensional features.

\subsubsection{Attention Matrix}

We evaluate a raw attention matrix feature baseline.
This feature is a direct, non-spectral variant of the LapEigvals feature, representing the raw self-attention maps without Laplacian eigendecomposition.
The parameter for the feature extraction is the same as LapEigvals.

\subsubsection{Word Embedding}

We extract the raw word embedding vectors directly from the embedding layer of the model.
These vectors represent the semantic embeddings of the LLM-generated tokens and do not contain positional information.
We collect the embedding vectors of all tokens in the generation.
We aggregate these vectors into a single sentence-level representation using the \texttt{mean} pooling operation.

\subsection{Competitors and Baseline Configurations}

We compare MMD-Flagger against two major classes of internal-state and output-level baselines.

We implement Kernel Language Entropy (KLE)~\citep{Nikitin-2024-kernel}, Lexical Similarity (LexSim)~\citep{Fomicheva-2020-unsupervised}, Eccentricity (Ecc)~\citep{Lin-2024-generating}, and EigenScore~\citep{Chen-2024-inside} using the official \texttt{LM-polygraph} package~\citep{Shelmanovvashurin-2025}\footnote{The source code for this implementation is obtained from \url{https://github.com/IINemo/lm-polygraph/releases/tag/v0.7.0}}.
For the fair comparison with MMD-Flagger, we feed the same population of stochastically generated responses to KLE, LexSim, Ecc, and EigenScore as MMD-Flagger.

\textbf{Kernel Language Entropy (KLE)} is a black-box estimation method that estimates the uncertainty of a given LLM's response (text).
Since it is the black-box method, a text-embedding model to convert a text into a fixed length vector is required.
Following the implementation in \texttt{LM-polygraph}, we use the \texttt{DeBERTa-large-MNLI} model for this purpose.\footnote{\url{https://huggingface.co/microsoft/deberta-large-mnli}}

\textbf{Lexical Similarity (LexSim)} is a black-box uncertainty estimation method that measures the surface lexical similarity among stochastically generated responses.
It estimates uncertainty by calculating the average pairwise lexical overlap across multiple generated responses.
In our implementation via the \texttt{LM-polygraph} package, we employ ROUGE-L as the default lexical similarity metric.

\textbf{Eccentricity (Ecc)} is a black-box uncertainty estimation method that constructs a similarity graph from stochastically generated responses.
It measures uncertainty by analyzing the geometric distance or deviation of each generated response from the centroid of all sampled responses in the similarity space.
Following the implementation in \texttt{LM-polygraph}, it leverages an external Natural Language Inference (NLI) model or text embedding model to compute pairwise response similarities.

\textbf{EigenScore} is a white-box uncertainty estimation method that measures the semantic stability of the internal hidden states of the LLM.
It extracts hidden state vectors from a chosen layer of the model across multiple stochastically generated responses.
It then estimates uncertainty by constructing a covariance matrix of these hidden states and analyzing its eigenvalue spectrum to assess the geometric spread and consistency in the representation space.

\section{Comparison of MMD-Flagger and KLE on TriviaQA: Error Analysis}

This appendix section provides a comparison between MMD-Flagger and KLE following the LLM generation setting from~\cite{Nikitin-2024-kernel}.

\subsection*{Experimental Settings}

% This evaluation target focuses on factual correctness, adhering strictly to the setup of Nikitin et al. (2024).

\paragraph{LLM Generation} TriviaQA consists of question and gold short answer pairs without pre-generated LLM outputs or human hallucination labels.
We use \texttt{mistralai/Mistral-7B-Instruct-v0.1} to generate responses given TriviaQA prompts.
We apply stochastic sampling with 10 samples per temperature~\mbox{$\tau \in \{0.1, 0.2, 0.3, 0.4, 0.5, 0.6, 0.7, 0.8, 0.9, 1.0\}$}, yielding $100$ stochastic samples per prompt.
Each generation uses controlled unique random seeds.

\paragraph{Hallucination Label Generation (LLM-as-a-judge)} We construct binary hallucination labels using \texttt{Llama-3-8B-Instruct} in an LLM-as-a-judge configuration.
The judge prompt requests a binary yes or no answer indicating whether the proposed answer matches the expected gold answer.

\paragraph{Estimators} We compare two uncertainty estimators: KLE and MMD-Flagger.
For KLE, we utilize all 100 stochastic samples across $10$ temperature parameters $\tau$, whereas the original implementation by Nikitin et al. (2024) restricted sampling to 10 samples at $\tau=1.0$.
For MMD-Flagger, we employ a linear kernel.
The evaluated feature representations with MMD-Flagger are AttentionMat (top-100), Hidden states, LapEigvals (top-100), and Word embedding.
% We score MMD trajectories using \texttt{quantile\_residual} and \texttt{tau\_min\_and\_tauarg\_min}.
To classify the calculated score as hallucinated or non-hallucinated, we set a thereshold value at the $75\%$ percentile of scores over all calculated scores for each estimator.

\subsection*{Benchmarking Results (ROC-AUC)}

MMD-Flagger underperforms KLE on the TriviaQA dataset across all evaluated feature representations and scoring methods.
Table~\ref{tab:roc_auc} lists the full experimental results extracted from the ROC-AUC evaluation sheet.

\begin{table*}[htbp]
\centering
\caption{
    ROC-AUC benchmarking results on TriviaQA for KLE, MMD-Flagger, and EigenScore.
}
\label{tab:roc_auc}
% \resizebox{\textwidth}{!}{%
\begin{tabular}{llll}
\toprule
\textbf{Algorithm} & \textbf{Feature Name} & \textbf{Aggregation} & \textbf{ROC-AUC} \\
\midrule
KLE & Text & -- & \textbf{.828} \\
\midrule
EigenScore & Hidden states & last & .794 \\
 & Hidden states & avg & .778 \\
\midrule
%  & Hidden states & last & .757 \\
%  & AttentionMat (top-100) & -- & .744 \\
%  & Word embedding & avg & .714 \\
%  & LapEigvals (top-100) & -- & .694 \\
%  & Hidden states & avg & .682 \\
MMD-Flagger & AttentionMat (top-100) & -- & .587 \\
 & Hidden states & avg & .521 \\
 & Word embedding & avg & .519 \\
 & LapEigvals (top-100) & -- & .512 \\
 & Hidden states & last & .495 \\
\bottomrule
\end{tabular}%
% This table version has the column of "Scoring method". I did not explain the scoring method in the main text. Thus, I omit the column for this moment.
% But I wanna add a section to explain the scoring method in the appendix. How can I justify the usage of the two scoring methods?
% \begin{tabular}{lllll}
% \toprule
% \textbf{Algorithm} & \textbf{Feature Name} & \textbf{Aggregation} & \textbf{Scoring Method} & \textbf{ROC-AUC} \\
% \midrule
% KLE & Text & -- & -- & \textbf{.828} \\
% \midrule
% EigenScore & Hidden states & last & -- & .794 \\
%  & Hidden states & avg & -- & .778 \\
% \midrule
%  & Hidden states & last & quantile\_residual & .757 \\
%  & AttentionMat (top-100) & -- & quantile\_residual & .744 \\
%  & Word embedding & avg & quantile\_residual & .714 \\
%  & LapEigvals (top-100) & -- & quantile\_residual & .694 \\
%  & Hidden states & avg & quantile\_residual & .682 \\
% MMD-Flagger & AttentionMat (top-100) & -- & tau\_min\_and\_tauarg\_min & .587 \\
%  & Hidden states & avg & tau\_min\_and\_tauarg\_min & .521 \\
%  & Word embedding & avg & tau\_min\_and\_tauarg\_min & .519 \\
%  & LapEigvals (top-100) & -- & tau\_min\_and\_tauarg\_min & .512 \\
%  & Hidden states & last & tau\_min\_and\_tauarg\_min & .495 \\
% \bottomrule
% \end{tabular}%
% }
\end{table*}

\subsubsection*{Error Analysis: Outcome Variations and Trajectory Dynamics}

We examine errors of estimators (MMD-Flagger and KLE) in the following two cases.

\begin{enumerate}
    \item Case 1: MMD-Flagger detects the hallucination correctly while KLE fails to detect it.
    \item Case 2: MMD-Flagger fails to detect the hallucination while KLE detects it correctly.
\end{enumerate}

\paragraph{Case 1: MMD-Flagger detects the hallucination correctly while KLE fails to detect it.}

\begin{table*}[htbp]
\centering
\caption{Prompt and hypothesis response for Case 1.}
\label{tab:case1_prompt}
\noindent\fbox{
\parbox{\dimexpr\textwidth-2\fboxsep-2\fboxrule}{\small
\textbf{Prompt:}\\
Answer the following question as briefly as possible.\\
Question: What was the name of the cat in Rising Damp?\\
Answer: vienna\\[0.3em]
Question: Which enduring cartoon character was created by Bob Clampett for the 1938 cartoon Porky's Hare Hunt?\\
Answer: bugs bunny\\[0.3em]
Question: The 1947 novel ``I, the Jury'', by New York author Mickey Spillane, was the first to feature which famous detective?\\
Answer: mike hammer\\[0.3em]
Question: How is his holiness Tenzin Gyatso better known?\\
Answer: dalai lama\\[0.3em]
Question: To which family of trees do junipers belong?\\
Answer: cypress\\[0.3em]
Question: What swirly pattern is named after the administrative central town of Renfrewshire in Scotland?\\
Answer:

\vspace{0.5em}
\textbf{Response with $\theta_{\rm hyp}$:} ``renfrewshire spiral''
}}
\end{table*}
\begin{table*}[htbp]
\centering
\caption{Selected stochastic generations for Case 1 across temperature parameters $\tau$.}
\label{tab:case1_generations}
\resizebox{\textwidth}{!}{%
\begin{tabular}{cl}
\toprule
\textbf{Temperature ($\tau$)} & \textbf{Generated LLM Samples (10 samples per $\tau$)} \\
\midrule
0.1 & renfrewshire spiral, renfrewshire spiral, renfrew, renfrew, renfrew, renfrew, renfrew, renfrewshire spiral, renfrew, renfrewshire spiral \\
0.3 & renfrew spiral, renfrewshire spiral, renfrew, renfrew, renfrew spiral, renfrew, renfrew spiral, renfrew, renfrew, renfrewshire spiral \\
0.5 & renfrew spiral, renfrewshire spiral, renfrew, renfrew, renal pattern, renfrew, renfrew spiral, renfrew, renfrew, renfrewshire spiral \\
0.8 & renfrew spiral, renfrewshire spiral, renfrew, renfrew, renfrew spiral, renfrew twist, renfrew spiral, renfrew, renfrew, renfrewshire \\
1.0 & neeps, renfrewshire spiral, renfrew, renfrew tower, re'nse, renfrew twist, renfrew spiral, renfrew, renfrew, mandorla \\
\bottomrule
\end{tabular}%
}
\end{table*}

\begin{figure}[htbp]
\centering
\includegraphics[width=0.5\textwidth]{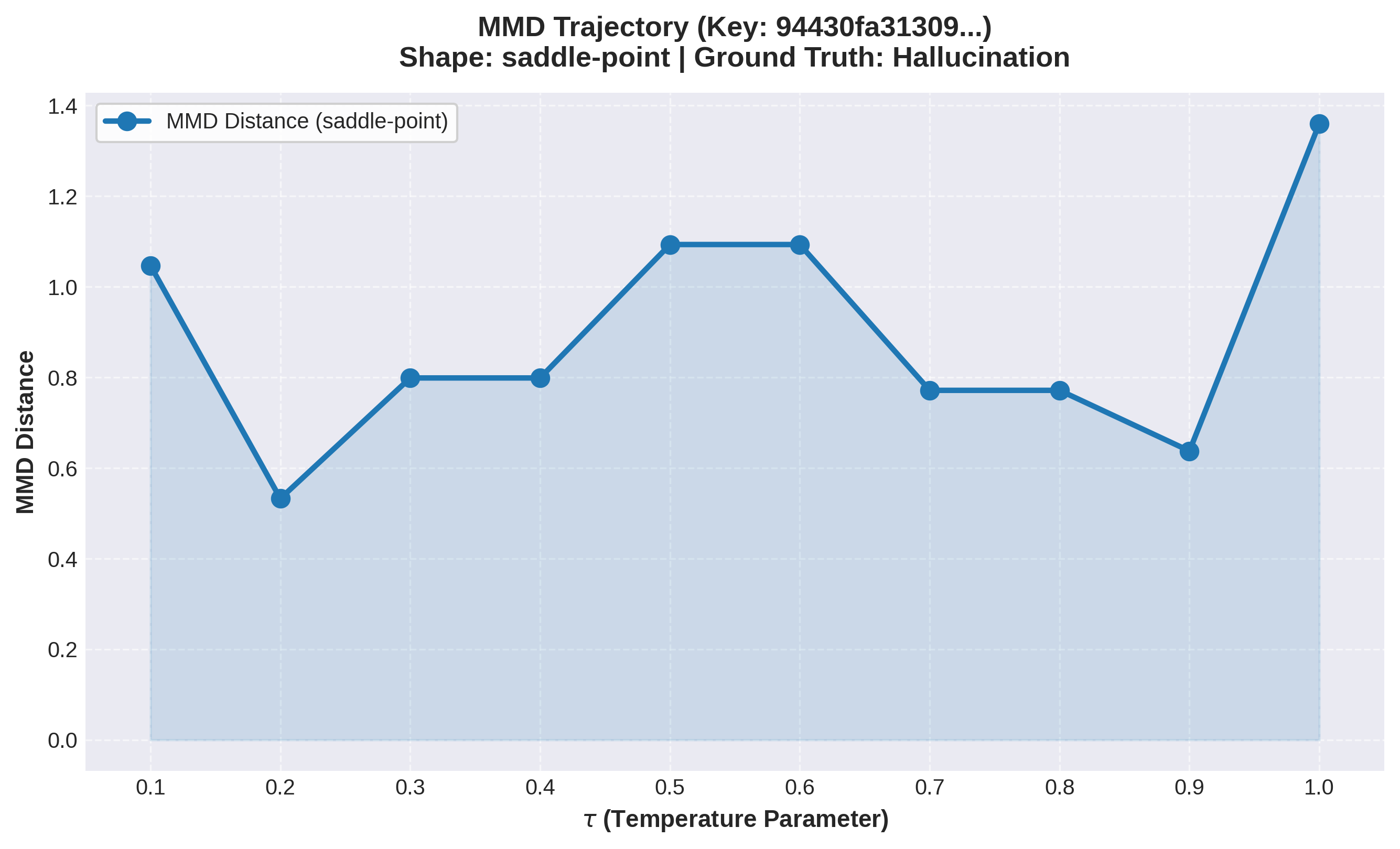}
\caption{
    MMD trajectory for Case 1 displaying a saddle-point shape. 
    MMD-Flagger correctly predicts hallucination, whereas KLE fails.}
\label{fig:case1}
\end{figure}

MMD-Flagger detects the hallucination, whereas KLE fails.
Table~\ref{tab:case1_prompt} shows the prompt and hypothesis response for Case 1, and Table~\ref{tab:case1_generations} details the stochastic LLM generations for Case 1 across representative temperatures $\tau$.
Figure~\ref{fig:case1} shows the MMD trajectory for Case 1.
Between $\tau=0.1$ and $\tau=0.4$, LLM generations remain constant (``renfrewshire spiral'' or ``renfrew''), followed by novel variations (``renal pattern'') at $\tau=0.5$.
MMD-Flagger captures this trajectory slope change, whereas KLE assigns low entropy due to text concentration.

\paragraph{Case 2: MMD-Flagger fails to detect the hallucination while KLE detects it correctly.}

\begin{table*}[htbp]
\centering
\caption{Prompt and hypothesis response for Case 2.}
\label{tab:case2_prompt}
\noindent\fbox{\parbox{\dimexpr\textwidth-2\fboxsep-2\fboxrule}{\small
\textbf{Prompt:}\\
Answer the following question as briefly as possible.\\
Question: What was the name of the cat in Rising Damp?\\
Answer: vienna\\[0.3em]
Question: Which enduring cartoon character was created by Bob Clampett for the 1938 cartoon Porky's Hare Hunt?\\
Answer: bugs bunny\\[0.3em]
Question: The 1947 novel ``I, the Jury'', by New York author Mickey Spillane, was the first to feature which famous detective?\\
Answer: mike hammer\\[0.3em]
Question: How is his holiness Tenzin Gyatso better known?\\
Answer: dalai lama\\[0.3em]
Question: To which family of trees do junipers belong?\\
Answer: cypress\\[0.3em]
Question: Which Glasgow group signed to Creation Records and recorded their debut single, ``All Fall Down'', in 1985?\\
Answer:

\vspace{0.5em}
\textbf{Response with $\theta_{\rm hyp}$:} ``the Jesus and Mary Chain''
}}
\end{table*}
\begin{table*}[htbp]
\centering
\caption{Stochastic generations for Case 2 across temperature parameters $\tau$.}
\label{tab:case2_generations}
\resizebox{\textwidth}{!}{%
\begin{tabular}{cl}
\toprule
\textbf{Temperature ($\tau$)} & \textbf{Generated LLM Samples (10 samples per $\tau$)} \\
\midrule
0.1 & the Jesus and Mary Chain (8 samples), the bunnymen (1 sample), the Jesus and Mary Chain \\
0.3 & the cocteau twins, the joy division, the Jesus and Mary Chain, the Cocteau Twins, the Jesus and Mary Chain, the Cocteau Twins, the bunnymen, the bunnymen, the Jesus and Mary Chain, cocteau twins \\
0.5 & the joy division, the joy division, primal scream, the Cocteau Twins, the Jesus and Mary Chain, cocteau twins, the bunnymen, the bunnymen, the bunnymen, cocteau twins \\
0.8 & the prime ministers, cocteau twins, primal scream, the Cocteau Twins, cocteau twins, cocteau twins, the bunnyside, the bunnymen, the sexual object, cocteau twins \\
1.0 & the prime ministers, goblins, primal scream, 1980sTeenage Fanclub, cocteau twins, cocteau twins, the coat, the bunnymen, the sexual object, cocteau twins \\
\bottomrule
\end{tabular}%
}
\end{table*}

\begin{figure}[htbp]
\centering
\includegraphics[width=0.4\textwidth]{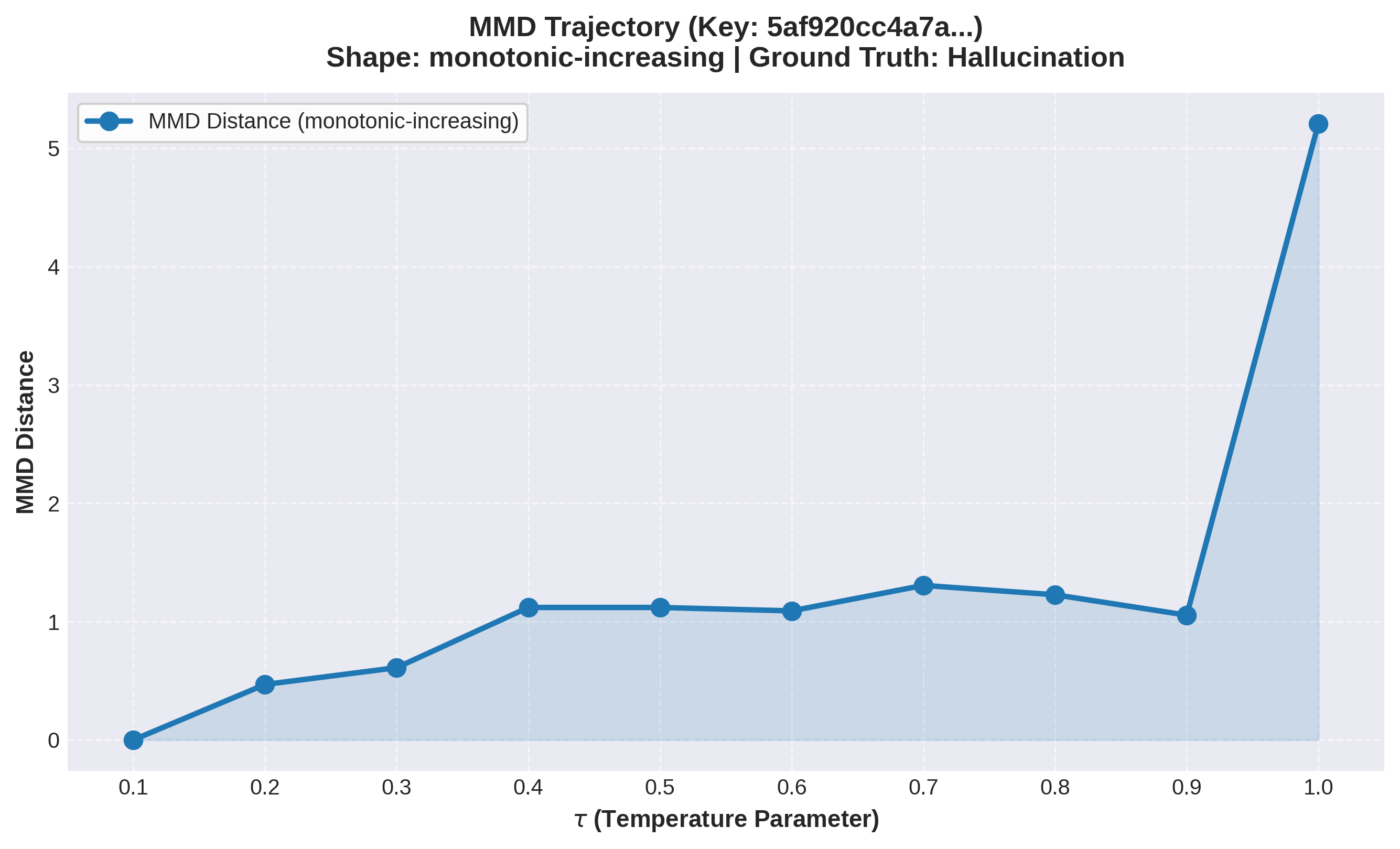}
\caption{
    MMD trajectory for Case 2 displaying a monotonic-increasing shape. 
    KLE correctly detects hallucination, while MMD-Flagger fails.}
\label{fig:case2}
\end{figure}

KLE detects the hallucination, whereas MMD-Flagger fails.
Table~\ref{tab:case2_prompt} shows the prompt and hypothesis response for Case 2, and Table~\ref{tab:case2_generations} details the stochastic LLM generations for Case 2 across representative temperatures $\tau$.
Figure~\ref{fig:case2} shows the MMD trajectory for Case 2.
At $\tau=0.1$, outputs are identical to the hypothesis response, but outputs gradually diverge into multiple distinct bands (``the cocteau twins'', ``the joy division'', ``primal scream'') as $\tau$ increases to 1.0.
This gradual divergence creates a monotonic increasing MMD trajectory that yields low scores.
% under \texttt{quantile\_residual} and \texttt{tau\_min\_and\_tauarg\_min}, whereas KLE detects high overall diversity.

\section{Discussion on Methodological Extensions and Generalizability}

\subsection*{Applicability to Alternative Sampling Strategies}
\label{subsec:alt-sampling}
% Address top-p, top-k sampling vs. temperature scaling here.

MMD-Flagger is designed to measure differences of LLM's generations by changing the decoding parameter $\theta$ from a narrower probability distribution of tokens to a broader probability distribution of tokens.
To realize this shift of probability distributions in the most straight-forward approach, we implement temperature scaling ($\tau$).

There are several alternative approaches to realize this shift.
The first alternative is, with a fixed $\tau$ parameter, changing the $k \in \mathbb{N}$ parameter in top-k sampling from a smaller $k$ to a larger $k$.
As the $k$ parameter becomes larger, the probability distribution of tokens becomes broader, which is similar to increasing the $\tau$ parameter.

The second alternative is, similar with first alternative, changing the parameter $p \in [0.0, 1.0]$ of top-p sampling from a smaller $p$ to a larger $p$.
As the $p$ parameter becomes larger, the probability distribution of tokens becomes broader, which is similar to increasing the $\tau$ parameter.

Combining the $k$ and $p$ parameters and changing both parameters from smaller to larger values is another approach to realize the shift. 

\subsection*{Alternative Representation Similarity Metrics}
\label{subsec:alt-metrics}
% Address replacing MMD with metrics like Centered Kernel Alignment (CKA) here.
The central idea of MMD-Flagger is to measure the difference between the representation distributions.
Therefore, other distance metrics that measures the distance between two probability distributions can be used instead of MMD.

For example, Centered Kernel Alignment (CKA)~\citep{Kornblith-2019-similarity} can be used instead of MMD.
The major difference lies in the target of measurement. 
While MMD whether two sets of samples ($S_{\rm hyp}, S_{\rm sto}^{\tau}$) come from the same underlying probability distribution, CKA measures whether two representations preserve the same geometry/relative relationships.
Thus, CKA can be used to measure the difference between two representations.

\section{Case study: one sample at the hypothesis generation}
\label{sec:case-one-sample-generation}

The MMD estimator in MMD-Flagger is defined in Eq.~\ref{eq:mmd-unbiased-est} under the setting that $S_{\text{hyp}}$ has $n$ samples.
However, there are cases where we only have one sample in $S_{\text{hyp}}$ (i.e., $n=1$).

We define below the MMD estimator for the case where~\mbox{$n=1$}.
This case is regarded as the kernel mean embedding of a Dirac delta distribution~\citep[Chapter 3]{Muandet-2017}.
A probability distribution $P$ is regarded as the Dirac delta~\mbox{$\delta_{\mathbf{h}_{\text{hyp}}^1}$} and $Q$ is the distribution of samples at temperature $\tau$.

\begin{equation} 
  \label{eq:mmd-unbiased-est-one-sample}
  \begin{split}
    \hatmmd^{2}_U & ( S_{\text{hyp}}, S_{\text{sto}}^{\tau} ) = \\
    & \frac{1}{m(m-1)} \sum_{1 \leq j \ne j' \leq m} k(\mathbf{h}_{\text{sto}}^{\tau, j}, \mathbf{h}_{\text{sto}}^{\tau, j'}) \\
    & - \frac{2}{m} \sum_{j=1}^m k(\mathbf{h}_{\text{hyp}}^{1}, \mathbf{h}_{\text{sto}}^{\tau, j}) + k(\mathbf{h}_{\text{hyp}}^{1}, \mathbf{h}_{\text{hyp}}^{1}).
  \end{split}
\end{equation}

\end{document}